\documentclass{article} 
\usepackage{iclr2027_conference,times}

\usepackage{amsmath,amsfonts,bm}

\def\eqref#1{equation~\ref{#1}}

\def\1{\bm{1}}

\DeclareMathAlphabet{\mathsfit}{\encodingdefault}{\sfdefault}{m}{sl}
\SetMathAlphabet{\mathsfit}{bold}{\encodingdefault}{\sfdefault}{bx}{n}

\usepackage{hyperref}
\usepackage{url}
\usepackage{amssymb}
\usepackage[table]{xcolor}
\usepackage{pifont}
\usepackage{multirow}
\usepackage{graphicx}
\usepackage{wrapfig}

\title{GraLoD: Graphics-Inspired Continuous Level-of-Detail Learning for Image Restoration}

\author{Hu Gao \& Lizhuang Ma \\
Department of Computer Science \\
Shanghai Jiao Tong University \\
Shanghai, China \\
\texttt{\{gao\_h, lzma\}@sjtu.edu.cn} \\
\And
Yulong Chen \\
 Department of Architecture and Design \\
Harbin Institute of Technology \\
Heilongjiang, China\\
\texttt{\{llong\_c\}@hit.edu.cn} \\
}

\iclrfinalcopy 
\begin{document}

\maketitle

\begin{abstract}
The spatial support required for image restoration varies across degradation types, image regions, and reconstruction stages. However, most existing methods rely on predefined multi-scale hierarchies and aggregate features through fixed fusion or attention, leaving the representation scale itself largely determined by the network architecture. This limitation becomes more pronounced when a task-specific backbone is extended to heterogeneous degradations in all-in-one restoration. Inspired by level-of-detail (LOD) rendering in computer graphics, we propose \textbf{GraLoD}, a plug-and-play framework that treats restoration scale as a spatially varying and stage-dependent continuous variable. GraLoD reuses the native encoder hierarchy, aligns its multi-scale features into a shared LOD representation space, and predicts a stage-conditioned LOD field at each decoder stage. Each spatial location then continuously queries only two neighboring representation levels, enabling the effective restoration scale to adapt to both local image content and reconstruction progress. To prevent degenerate or arbitrary scale selection, we further introduce minimal-sufficient footprint calibration (MSFC) together with structure-aware regularization (SAR) to encourage restoration-effective and spatially coherent LOD assignments. GraLoD can be directly integrated into existing restoration backbones without redesigning their fundamental feature-processing blocks. Extensive experiments demonstrate consistent improvements in task-specific and all-in-one restoration.
\end{abstract}

\section{Introduction}
\label{sec:introduction}

Image restoration aims to recover high-quality images from observations degraded by corruptions. Recent CNN-, Transformer-, and state-space-based restorers have achieved strong performance by combining local feature extraction with increasingly effective long-range modeling~\cite{starir11429607,lin2026derestormer,ALGgao2024learning}. Most of these methods adopt hierarchical encoder--decoder architectures, in which high-resolution features preserve local details while lower-resolution features provide progressively broader contextual information.

However, the spatial context required for restoration is inherently non-uniform. Large or spatially extended degradations often require broader contextual reasoning, whereas fine textures, object boundaries, and thin structures depend more strongly on high-resolution representations. Such variation also occurs within the same image. Nevertheless, most restoration models typically use a predefined set of feature scales and combine them through skip connections, feature fusion, or attention. Although these mechanisms can adapt the contribution of different levels, the underlying representation scales remain discrete and fixed by the architecture. As a result, the model mainly learns how to combine predefined scales rather than which scale is most appropriate for each spatial location.

This limitation becomes more evident in all-in-one image restoration, where a single model must accommodate heterogeneous degradations with substantially different spatial characteristics. Several approaches improve such flexibility through degradation representations, prompts, or adaptive conditioning~\cite{clearAIR,Allrestor11367271,CAPTNet10526271}. Scale-adaptive designs further adjust receptive fields or sampling locations according to degradation patterns~\cite{he2025scaleadaptive}. These strategies improve degradation adaptation and spatial aggregation, but the resolution of the queried representation is still largely constrained by the predefined network hierarchy.

We therefore argue that restoration scale itself should be treated as a spatially varying variable. This perspective is closely related to level-of-detail (LOD) rendering in computer graphics. Classical mipmapping selects an appropriate representation resolution according to the spatial footprint of a rendered sample~\cite{williams1983pyramidal}, and similar scale-aware principles have been adopted in neural rendering~\cite{barron2021mipnerf}. The underlying principle is that different samples should access representations at different resolutions according to the spatial support they require. Unlike rendering, however, image restoration does not have an explicitly known sampling footprint, the appropriate restoration scale must instead be inferred from the degraded observation and the current reconstruction state.

Inspired by this principle, we propose \textbf{GraLoD}, a plug-and-play framework that introduces continuous LOD adaptation into existing restoration backbones. Rather than constructing an additional feature pyramid, GraLoD directly reuses the native multi-scale features produced by the encoder and aligns them into a shared LOD representation space. At each decoder stage, a stage-conditioned LOD field is predicted according to both the current reconstruction state and the aligned encoder hierarchy. Each spatial location then continuously queries only two neighboring LOD levels, allowing the effective restoration scale to adapt not only across image regions but also throughout the coarse-to-fine reconstruction process. The queried representation is injected through a lightweight residual adaptor, while the original encoder, decoder, and skip connections remain unchanged.
To prevent the learned LOD field from degenerating into arbitrary multi-scale gating or collapsing toward a preferred level, we further introduce minimal-sufficient footprint calibration (MSFC). MSFC evaluates the local restoration utility of different representation levels and encourages each location to use the smallest spatial footprint sufficient for accurate reconstruction. In addition, structure-aware regularization (SAR) promotes coherent LOD assignments within homogeneous regions while allowing scale transitions around structural boundaries. 

\begin{wrapfigure}{l}{0.66\textwidth}
    \centering
  
    \includegraphics[width=\linewidth]{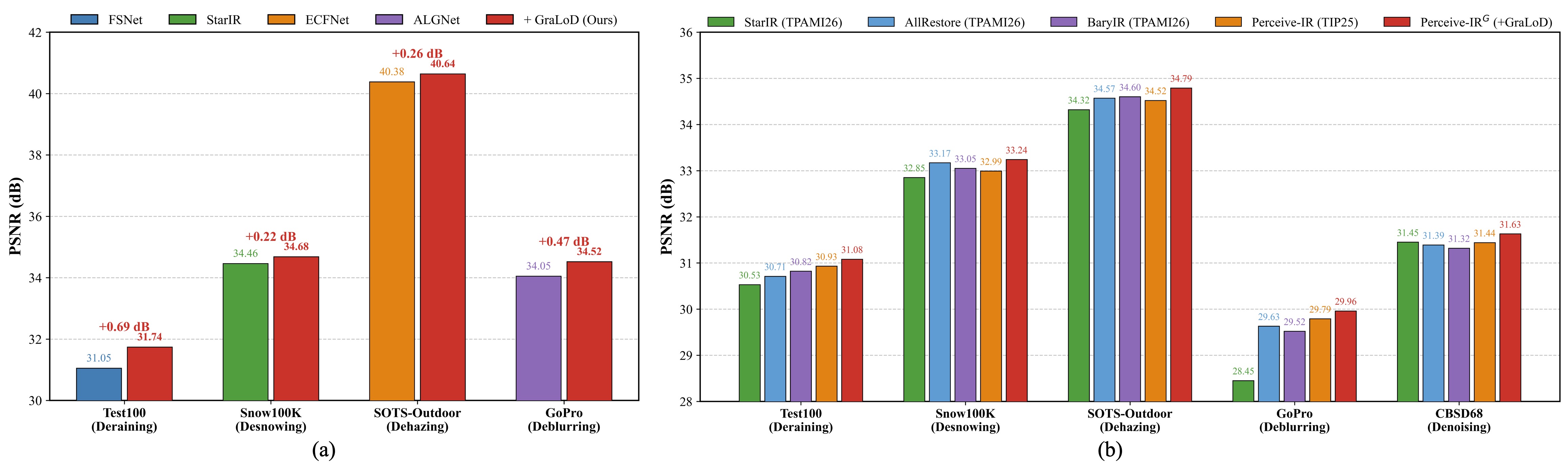}
    \caption{
    Effectiveness of GraLoD in different restoration settings.
    (a) GraLoD consistently improves representative task-specific restoration backbones.
    (b) GraLoD further improves an all-in-one restorer and achieves strong performance across five degradation types.
    }
    \label{fig:intro_performance}

\end{wrapfigure}

GraLoD can be directly integrated into existing restoration models without redesigning their fundamental feature-processing blocks. Under task-specific training, it improves restoration through spatially and stage-wise adaptive scale selection, while under multi-degradation training, the same mechanism enables existing backbones to accommodate heterogeneous restoration requirements and extend toward all-in-one restoration. As shown in Fig.~\ref{fig:intro_performance}, GraLoD consistently improves representative task-specific backbones by up to $0.69$ dB and further enhances unified restorers across five degradation types, achieving competitive or superior performance over existing all-in-one methods.

Our contributions are summarized as follows:
\begin{itemize}
    \item We introduce a graphics-inspired formulation that treats restoration scale as a spatially varying and stage-dependent continuous representation variable rather than a predefined architectural choice.
    
    \item We propose \textbf{GraLoD}, a plug-and-play continuous LOD framework that reuses the native encoder hierarchy, aligns it into a shared LOD space, and performs stage-conditioned continuous scale querying without redesigning the underlying restoration backbone.
    
    \item We introduce minimal-sufficient footprint calibration (MSFC) and structure-aware regularization (SAR) to constrain LOD learning toward restoration-effective and spatially meaningful scale assignments.
    
    \item Extensive experiments demonstrate that GraLoD consistently improves existing backbones in task-specific and all-in-one restoration settings and generalizes effectively to diverse degradation conditions.
\end{itemize}

\section{Related Work}
\label{sec:related_work}

\subsection{Image Restoration}
\label{sec:related_ir}

Image restoration has evolved from prior-driven formulations toward data-driven representation learning. Traditional methods rely on explicit structural assumptions or degradation models to constrain the inverse problem~\cite{10558778,11342300}, whereas recent CNN-, Transformer-, diffusion-, and Mamba-based approaches learn restoration priors directly from data~\cite{gao2025mixed,lin2026derestormer,UniUIR}. Task-specific methods have improved reconstruction through local--global feature interaction, spatial--frequency modeling, predictive filtering, and task-dependent priors. ALGNet~\cite{ALGgao2024learning} combines adaptive local and global representations, XYScanNet~\cite{liu2024xyscannet} enhances long-range dependency modeling through alternating spatial scans, and EfDeRain+~\cite{efderainguo2025efficientderain+} formulates deraining as a predictive filtering process. Other studies exploit frequency-domain representations~\cite{FDTANetgao2025frequency}, explicit image priors~\cite{PGH2Netisu2025prior}, or generative diffusion models~\cite{upid10.1145/3664647.3680560,diffunmix10656884}. Although these designs substantially improve task-specific restoration, their feature hierarchies and scale configurations are generally fixed once the network architecture is determined, limiting their ability to adapt the effective restoration scale to spatially varying degradation patterns and image structures.

All-in-one image restoration aims to further handle multiple degradations within a single model. Existing methods mainly improve degradation perception, representation disentanglement, or conditional feature adaptation. AdaIR~\cite{cui2025adair} separates degradation-related and content-related information, Perceive-IR~\cite{Perceive-IR10990319} models both degradation category and severity, and BaryIR~\cite{baryir11417902} aligns heterogeneous degradation distributions through a shared representation. Other approaches enhance unified restoration through long-range interaction, multimodal conditioning, or vision-language priors~\cite{starir11429607,Allrestor11367271,VLUNetZeng_2025_CVPR}.

These methods improve adaptation across degradation types, yet most still rely on predefined multi-scale hierarchies and mainly adjust degradation representations, prompts, routing, or feature modulation. As a result, degradations with different spatial extents are still handled within the same discrete set of backbone resolutions, while representation scale itself is seldom treated as a spatially varying variable. GraLoD targets this limitation by introducing continuous representation-scale adaptation into restoration backbones. It supports both task-specific and all-in-one restoration without altering the backbone's fundamental feature-processing blocks.

\subsection{Scale Adaptation and Level-of-Detail Representations}
\label{sec:related_scale}

Multi-scale representations are widely used in image restoration to combine fine spatial details with broader contextual information. Encoder--decoder architectures typically form hierarchical features through progressive downsampling. Pyramid Attention~\cite{mei2023pyramid} extends non-local matching to multiple scales, while EPA-Net~\cite{hua2026epa} dynamically weights pyramid features. State-space and attention-based restorers also combine local and global representations to improve reconstruction.
Spatial aggregation can also be adapted according to degradation characteristics. DeRestormer~\cite{lin2026derestormer} combines deformable attention with multi-scale aggregation for localized degradations, while DSASformer~\cite{wang2026dsasformer} employs dynamic scale-aware sparse attention together with multi-scale detail enhancement. These methods provide greater flexibility in receptive fields, sampling locations, and cross-scale feature interaction, but the underlying representation levels remain predefined by the network hierarchy.
Level-of-detail adaptation has long been studied in computer graphics. Classical mipmapping selects progressively filtered representations according to the projected sampling footprint~\cite{williams1983pyramidal}. Mip-NeRF~\cite{barron2021mipnerf} extends this idea to scale-aware neural rendering, while Mip-Splatting~\cite{yu2024mipsplatting} incorporates frequency-aware filtering for alias-free Gaussian rendering. Continuous LOD representations have also been explored to support smooth resolution changes within unified 3D scene representations~\cite{milef2025clod,cheng2026clodgs}.

GraLoD brings this idea to image restoration by treating the representation scale required for local reconstruction as a spatially varying variable. Rather than selecting from fixed scales independently, it predicts a continuous LOD coordinate and queries neighboring levels in an ordered LOD representation space, while remaining compatible with existing restoration backbones.

\section{Method}
\label{sec:method}

\subsection{Problem Formulation}
\label{sec:problem}

Given a degraded image $\mathbf{y}\in\mathbb{R}^{H\times W\times3}$,
most models adopt an encoder--decoder architecture and naturally produce hierarchical features $F^l=\mathcal{E}_l(\mathbf{y})$, $l=0,\ldots,L-1$, where $F^l$ is ordered from fine to coarse resolution.
Although these features provide different spatial supports, their usage is typically determined by predefined skip connections or multi-scale fusion.
We instead treat restoration scale as a spatially varying continuous variable.
For decoder stage $s$, we introduce an LOD field:
\begin{equation}
    \boldsymbol{\lambda}_s
    =
    \{\lambda_s(\mathbf{p})\mid\mathbf{p}\in\Omega_s\},
    \qquad
    \lambda_s(\mathbf{p})\in[0,L-1],
    \label{eq:stage_lod}
\end{equation}
where $\Omega_s$ denotes the spatial lattice of stage $s$.
$\lambda_s(\mathbf{p})$ describes the representation scale required to reconstruct location $\mathbf{p}$ at the current decoding stage.
A smaller value emphasizes fine spatial evidence, whereas a larger value accesses representations with broader contextual support.
In this paper, our GraLoD reuses the native encoder hierarchy and converts the discrete backbone scales into a continuously queryable LOD space.

\begin{figure*}[t]
    \centering
    \includegraphics[width=\linewidth]{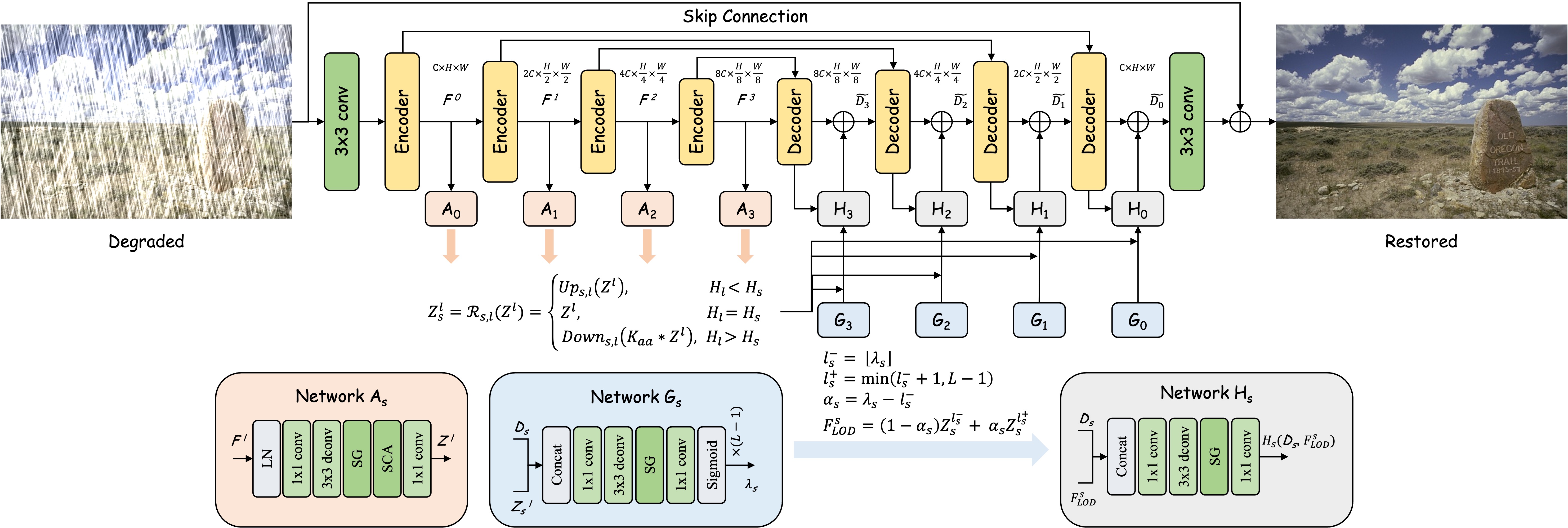}
    \caption{
    Architecture of GraLoD.
    The native encoder hierarchy is aligned by $\mathcal{A}_l$ into an ordered LOD representation space and shared across decoder stages.
    At stage $s$, the aligned features are resized by $\mathcal{R}_{s,l}$ and combined with $D_s$ to predict the spatial LOD field through $\mathcal{G}_s$.
    Two neighboring LOD levels are then continuously interpolated to obtain $F_{\mathrm{LOD}}^s$, which is injected into the decoder through the residual adaptor $\mathcal{H}_s$.
    The detailed structures of $\mathcal{A}_l$, $\mathcal{G}_s$, and $\mathcal{H}_s$ are shown at the bottom.
    }
    \label{fig:framework}
\end{figure*}

\subsection{GraLoD: Plug-and-Play Continuous LOD Adaptation}
\label{sec:gralod}

Figure~\ref{fig:framework} illustrates how GraLoD is integrated into a conventional encoder--decoder backbone. The native multi-scale encoder features are first aligned into a shared LOD representation space. At each decoder stage, GraLoD predicts a stage-specific spatial LOD field, queries the corresponding representation scale, and injects the resulting feature through a lightweight residual adaptor. In this way, the original encoder, decoder, and skip connections are fully preserved, while GraLoD only augments representation-scale selection.
Encoder features from different stages vary in channel dimensionality, spatial resolution, and representation characteristics, making direct interpolation between them unsuitable. We therefore map each encoder feature into a shared space:
\begin{equation}
Z^l=\mathcal{A}_l(F^l),
\qquad l=0,\ldots,L-1,
\label{eq:lod_alignment}
\end{equation}
where $\mathcal{A}_l$ is a lightweight local refinement. The aligned hierarchy
$\mathcal{Z}={Z^0,Z^1,\ldots,Z^{L-1}}$
forms an ordered LOD representation space, in which finer levels preserve detailed spatial information while deeper levels provide progressively broader contextual support. When resolution conversion is required, anti-aliased resampling is applied to reduce cross-scale inconsistency. This aligned hierarchy is constructed only once and shared by all decoder stages.
Let $D_s$ denote the feature entering decoder stage $s$. To match the spatial resolution of the current decoding stage, each $Z^l$ is aligned to $\Omega_s$:
\begin{equation} 
Z_s^l=\mathcal{R}_{s,l}(Z^l), 
\qquad l=0,\ldots,L-1, 
\label{eq:stage_alignment} 
\end{equation}
where $\mathcal{R}_{s,l}$ denotes scale alignment. A lightweight estimator $\mathcal{G}_s$ then predicts the LOD field:
\begin{equation}
\boldsymbol{\lambda}_s =\mathcal{G}_s(D_s,Z_s^0,\ldots,Z_s^{L-1}),
\label{eq:stage_lod_estimation}
\end{equation}
where each element $\lambda_s(\mathbf{p})$ of
$\boldsymbol{\lambda}_s$ represents the predicted LOD coordinate at location
$\mathbf{p}$. Conditioning scale selection on $D_s$ allows the requested representation scale to depend jointly on spatial content and reconstruction progress. 
Given $\lambda_s(\mathbf{p})$, its two adjacent LOD levels are defined as $l_s^-(\mathbf{p}) = \left\lfloor\lambda_s(\mathbf{p})\right\rfloor$, $l_s^+(\mathbf{p})
= \min \left( l_s^-(\mathbf{p})+1,L-1 \right)$, with interpolation coefficient
$\alpha_s(\mathbf{p}) = \lambda_s(\mathbf{p}) - l_s^-(\mathbf{p})$. The continuously queried representation is then computed as:
\begin{equation}
F_{\mathrm{LOD}}^s(\mathbf{p})
=
(1-\alpha_s(\mathbf{p}))
Z_s^{\,l_s^-(\mathbf{p})}(\mathbf{p})
+
\alpha_s(\mathbf{p})
Z_s^{\,l_s^+(\mathbf{p})}(\mathbf{p}).
\label{eq:continuous_lod}
\end{equation}

Unlike generic scale attention, which assigns independent weights to all
feature levels, GraLoD represents scale with a single continuous coordinate
along an ordered LOD axis and interpolates only between two neighboring
levels. This formulation preserves the ordering of the representation scales
and allows the queried scale to vary continuously across spatial locations
and decoder stages. The continuity and gradient properties of the LOD query,
together with the motivation for stage-conditioned prediction, are discussed
in Appendix~\ref{app:continuous_lod}.
Finally, the queried representation is injected into the original decoder
through a lightweight residual adaptor:
\begin{equation}
\widetilde{D}_s
= D_s + \mathcal{H}_s(D_s,F_{\mathrm{LOD}}^s),
\label{eq:stage_adapter}
\end{equation}
where $\mathcal{H}_s$ denotes the adaptor at decoder stage $s$.
The enhanced feature $\widetilde{D}_s$ is then processed by the original
decoder stage together with its native skip connection. The aligned LOD
hierarchy $\mathcal{Z}$ is shared across decoder stages, whereas the LOD
estimator $\mathcal{G}_s$ and residual adaptor $\mathcal{H}_s$ are
stage-specific. GraLoD therefore introduces representation-scale adaptation
throughout decoding while preserving the original encoder, decoder, and skip
connections. More details are provided in Appendix~\ref{app:integration}.

\subsection{LOD Calibration and Structural Regularization}
\label{sec:lod_regularization}

Although the reconstruction loss can guide the LOD field, it does not uniquely
determine the appropriate representation scale. Different LOD assignments may
produce similar reconstruction errors, causing the predicted coordinates to
collapse toward a preferred level or behave as unconstrained multi-scale
gating. We therefore introduce minimal-sufficient footprint calibration (MSFC),
which encourages each location to use the smallest representation footprint
that provides sufficient restoration accuracy.
During training, a lightweight probe $\mathcal{P}_s$, shared across LOD levels
within decoder stage $s$, evaluates each aligned representation as
$\hat{\mathbf{x}}_s^{\,l}
=\mathcal{P}_s(\operatorname{Concat}(D_s,Z_s^l))$.
Let $\mathbf{x}_s$ denote the clean target resized to the spatial resolution of
stage $s$. The local restoration error is measured by:
\begin{equation}
e_{s,l}(\mathbf{p})
=\mathcal{A}_r(|\hat{\mathbf{x}}_s^{\,l}-\mathbf{x}_s|)(\mathbf{p}),
\label{eq:calibration0}
\end{equation}
where $\mathcal{A}_r$ denotes local averaging. To discourage unnecessarily
coarse representations, we define the scale cost as:
\begin{equation}
c_{s,l}(\mathbf{p})
=e_{s,l}(\mathbf{p})+\gamma l/(L-1).
\label{eq:calibration01}
\end{equation}
A soft distribution over LOD levels is then obtained by
$q_{s,l}(\mathbf{p})
\propto\exp(-c_{s,l}(\mathbf{p})/\tau)$, and the corresponding target coordinate
is $\lambda_s^*(\mathbf{p})
=\sum_{l=0}^{L-1}l\,q_{s,l}(\mathbf{p})$.
The predicted LOD field is calibrated toward this target through:
\begin{equation}
\mathcal{L}_{\mathrm{cal}}
=
\frac{1}{S}
\sum_{s=1}^{S}
\frac{1}{|\Omega_s|}
\sum_{\mathbf{p}\in\Omega_s}
\left|
\lambda_s(\mathbf{p})
-
\operatorname{sg}
\left(
\lambda_s^*(\mathbf{p})
\right)
\right|,
\label{eq:calibration}
\end{equation}
where $\operatorname{sg}(\cdot)$ denotes stop-gradient. A coarser level is
therefore favored only when the additional spatial support provides sufficient
reconstruction benefit to compensate for its scale penalty. The probes are
used only during training and are removed at inference. Further details of
MSFC and probe training are provided in Appendix~\ref{app:msfc}.
We further regularize the spatial organization of the LOD field using the
clean image structure. For neighboring locations $\mathbf{p}$ and $\mathbf{q}$,
their affinity is defined as
$w_{\mathbf{p}\mathbf{q}}^s
=\exp(-\beta\|\mathbf{x}_s(\mathbf{p})
-\mathbf{x}_s(\mathbf{q})\|_1)$.
The structure-aware regularization (SAR) is:
\begin{equation}
\mathcal{L}_{\mathrm{str}}
=
\frac{1}{S}
\sum_{s=1}^{S}
\frac{1}{|\Omega_s|}
\sum_{\mathbf{p}\in\Omega_s}
\sum_{\mathbf{q}\in\mathcal{N}(\mathbf{p})}
w_{\mathbf{p}\mathbf{q}}^s
\left|
\lambda_s(\mathbf{p})
-
\lambda_s(\mathbf{q})
\right|.
\label{eq:structure_reg}
\end{equation}
This objective encourages similar LOD coordinates within homogeneous regions
while allowing scale changes across structural boundaries. Further analysis
is given in Appendix~\ref{app:structure}.
The overall training objective is:
\begin{equation}
\mathcal{L}
=
\mathcal{L}_{\mathrm{rec}}
+
\lambda_{\mathrm{cal}}\mathcal{L}_{\mathrm{cal}}
+
\lambda_{\mathrm{str}}\mathcal{L}_{\mathrm{str}}
+
\lambda_{\mathrm{probe}}\mathcal{L}_{\mathrm{probe}},
\label{eq:total_objective}
\end{equation}
where $\mathcal{L}_{\mathrm{rec}}$ follows the original objective of the
underlying restoration backbone and $\mathcal{L}_{\mathrm{probe}}$ supervises
the training-only scale probes. Additional training details and computational
analysis are provided in Appendices~\ref{app:training}
and~\ref{app:complexity}.

\section{Experiments}
\label{sec:exp}
In this section, we first present the comparison results, followed by the ablation studies. The experimental setup and more experiments are provided in Appendix~\ref{sec:expsetup} and~\ref{sec:expset}.

\begin{table*}
\centering
\caption{Quantitative comparison of GraLoD-enhanced image restoration models across different degradation types. $^{G}$ denotes the integration of GraLoD into the corresponding backbone.}
\label{tb:task_specific}
    \resizebox{\linewidth}{!}{
\begin{tabular}{ccc|ccc}
    \hline
     \multirow{2}{*}{Methods} & \multicolumn{2}{c|}{Deraining} & \multirow{2}{*}{Methods} & \multicolumn{2}{c}{Desnowing}
    \\
    &PSNR $\uparrow$ &SSIM $\uparrow$  &&PSNR $\uparrow$ &SSIM $\uparrow$ 
    \\
    \hline\hline
   EfDeRain+~\cite{efderainguo2025efficientderain+}  &31.10&0.911&  MSP-Former~\cite{mspformer10095605}  &33.43&\textbf{0.96} 
    \\
    MHNet~\cite{gao2025mixed} &31.25&0.901  &PEUNet~\cite{PEUNet10830558} &34.11&\textbf{0.96} 
     \\
     PPTformer~\cite{pptformerwang2025intra} & 31.48&\textbf{0.922}  &ECFNet~\cite{gao2026emphasizing} &34.26&\textbf{0.96} 
     \\
 ACL~\cite{aclgu2025acl} &\underline{31.51}&0.914 &PW-FNet~\cite{pwfnet11433521} &\underline{34.50}&\underline{0.95} 
     \\
     FSNet~\cite{FSNet} &31.05&0.919   &StarIR~\cite{starir11429607} &34.46&\textbf{0.96} 
    \\
     \rowcolor{gray!20}\textbf{FSNet$^G$}~\cite{FSNet} &\textbf{31.74}&\underline{0.921} & \textbf{StarIR$^G$}~\cite{starir11429607} &\textbf{34.68}&\textbf{0.96} 
     \\
     \hline
     \multirow{2}{*}{Methods} & \multicolumn{2}{c|}{Dehazing}& \multirow{2}{*}{Methods} & \multicolumn{2}{c}{Deblurring}
     \\
      &PSNR $\uparrow$ &SSIM $\uparrow$ &&PSNR $\uparrow$ &SSIM $\uparrow$
      \\
      \hline \hline
      FDTANet~\cite{FDTANetgao2025frequency}  &34.73&0.989 &  FSNet~\citep{FSNet} &33.29&0.963
      \\
      PGH$^2$Net~\cite{PGH2Netisu2025prior}&37.52&0.989 &LCDNet~\cite{gao2025enhancing} & 33.55 &	0.966
      \\
      StarIR~\cite{starir11429607}&38.98&0.991 & MBMamba~\cite{gao2025mbmamba} &33.89 &0.968
      \\
      Defusion~\cite{DefusionLuo_2025_CVPR} & 37.41&\underline{0.993} & StarIR~\cite{starir11429607}&\underline{34.34} &\textbf{0.970}
      \\
     ECFNet~\cite{gao2026emphasizing} &\underline{40.38}&\underline{0.993}   &ALGNet~\cite{ALGgao2024learning} &34.05 &\underline{0.969}
      \\
       \rowcolor{gray!20}\textbf{ECFNet$^G$}~\cite{gao2026emphasizing} &\textbf{40.64}&\textbf{0.994} & \textbf{ALGNet$^G$}~\cite{ALGgao2024learning} & \textbf{34.52} & \textbf{0.970}
    \\
    \hline
\end{tabular}}
\end{table*}

\begin{figure*} 
    \centerline{\includegraphics[width=1\linewidth]{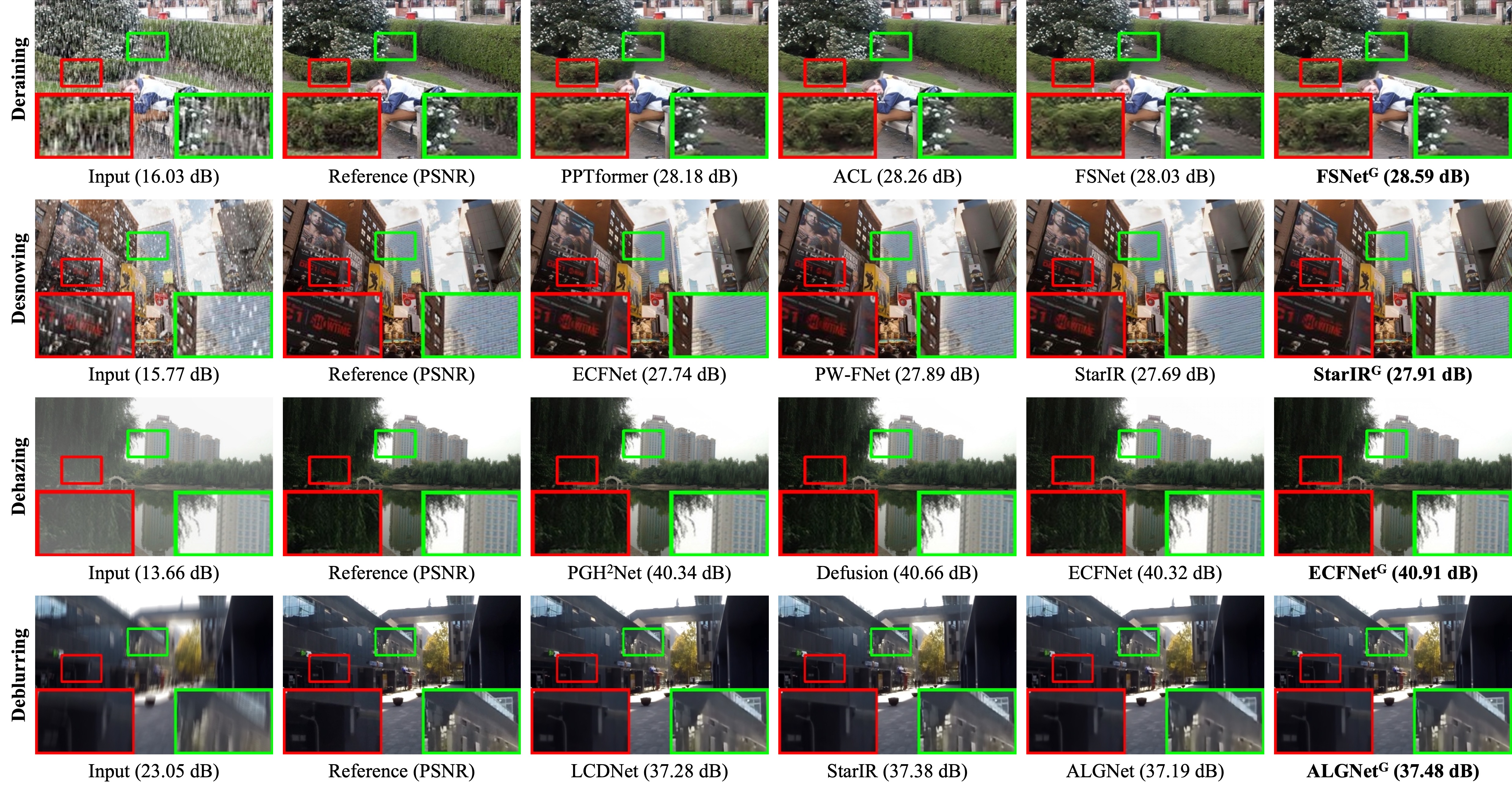}}
	\caption{Qualitative comparison in the task-specific setting.}
 \label{fig:qualitativesig}
\end{figure*}

\subsection{Results}

\label{sec:quantitative_results}

\subsubsection{Performance Improvement on Task-Specific Restoration}
\label{sec:task_specific_results}

We first evaluate whether GraLoD can improve existing restoration models without changing their original network. As shown in Table~\ref{tb:task_specific}, integrating GraLoD consistently improves representative restoration backbones across different degradation types. In deraining, GraLoD improves FSNet~\cite{FSNet} by 0.69 dB, allowing it to outperform the compared task-specific methods. Similar improvements are observed for desnowing, dehazing and deblurring, where StarIR~\cite{starir11429607}, ECFNet~\cite{gao2026emphasizing} and ALGNet~\cite{ALGgao2024learning} gain 0.22 dB, 0.26 dB and 0.47 dB, respectively, while their structural similarity is also maintained or slightly improved.
These results demonstrate that GraLoD reuses the native multi-scale hierarchy and adaptively determines the representation scale required at different spatial locations and reconstruction stages. The consistent gains support our motivation that even well-designed  restorers can benefit from explicitly adapting the representation scale during reconstruction. Importantly, these improvements are obtained while preserving the original backbone architecture, highlighting the plug-and-play nature of GraLoD.

Figure~\ref{fig:qualitativesig} presents qualitative comparisons of GraLoD-enhanced task-specific restoration models across different degradation types. Consistent with the quantitative results, integrating GraLoD improves the visual quality of the restored images. Compared with the corresponding baseline models, the GraLoD-enhanced variants produce cleaner restoration results with improved structural fidelity and fewer degradation residuals.

\begin{table*}
\centering
\caption{Quantitative comparison in the all-in-one setting across CNN-, Transformer-, and Mamba-based backbones. Results include task-specific (${\color{red}\blacklozenge}$), task-aligned (${\color{green}\blacklozenge}$), and all-in-one (${\color{blue}\blacklozenge}$) methods. }
\label{tb:allinone}
    \resizebox{\linewidth}{!}{
\begin{tabular}{c|cccccccccc|cc}
    \hline
    \multirow{2}{*}{Methods} & \multicolumn{2}{c}{Deraining} & \multicolumn{2}{c}{Desnowing} & \multicolumn{2}{c}{Dehazing}& \multicolumn{2}{c}{Deblurring}& \multicolumn{2}{c|}{Denosing}  & \multicolumn{2}{c}{Average} 
    \\
    &PSNR $\uparrow$ &  SSIM $\uparrow$  &PSNR $\uparrow$ &SSIM $\uparrow$ & PSNR $\uparrow$&SSIM $\uparrow$ &PSNR $\uparrow$ & SSIM $\uparrow$&PSNR $\uparrow$ & SSIM $\uparrow$&PSNR $\uparrow$ & SSIM $\uparrow$
    \\
    \hline\hline
   ${\color{red}\blacklozenge}$EfDeRain+~\cite{efderainguo2025efficientderain+} & 30.89 & \underline{0.909} &  29.42 & 0.925 & 33.01 & 0.952 & 26.48 & 0.804 & 30.62 & 0.880 & 30.08 & 0.894
     \\
     ${\color{red}\blacklozenge}$PGH$^2$Net~\cite{PGH2Netisu2025prior}& 29.52 & 0.896 & 31.88 & 0.932 & 34.34 & 0.976 & 25.96 & 0.789 & 30.97 & 0.882 & 30.53 & 0.895
     \\
     ${\color{red}\blacklozenge}$ALGNet~\cite{ALGgao2024learning}& 28.99 &0.896 & 30.89 & 0.924 & 32.76 & 0.968 & 29.41 & 0.882 & 30.97 & 0.881 & 30.60 & 0.910
     \\
       \rowcolor{gray!20} \textbf{ALGNet$^G$}~\cite{ALGgao2024learning} &30.71 & 0.901 & 32.92 & 0.946 & 34.09 & 0.978 & 29.59 & 0.881 & 31.21 & 0.883 & 31.70 & 0.918
     \\
     \hline
   ${\color{green}\blacklozenge}$ECFNet~\cite{gao2026emphasizing} & 30.44 & 0.890 & 32.54 & 0.932 & 33.99 & 0.978 & 27.58 & 0.836 & 31.10 & 0.882 & 31.13 & 0.904
    \\
     ${\color{green}\blacklozenge}$MHNet~\cite{gao2025mixed} & 30.26 & 0.888 & 31.94 & 0.941 & 33.98 & 0.976 & 27.41 & 0.830 & 31.33 & 0.882 & 30.98 & 0.903
     \\
     ${\color{green}\blacklozenge}$StarIR~\cite{starir11429607} &30.53 &0.891 & 32.85& 0.950 & 34.32&0.978 & 28.45 &0.880& \underline{31.45} & 0.885 & 31.52 & 0.917
     \\
       \rowcolor{gray!20} \textbf{StarIR$^G$}~\cite{starir11429607}& 30.81 & \underline{0.905} & \underline{33.23} &\textbf{0.952} & 34.59 & 0.979& 29.42 & 0.883 & 31.44 & \underline{0.886} & 31.90 & \underline{0.921}
     \\
     \hline
       ${\color{blue}\blacklozenge}$AllRestorer~\cite{Allrestor11367271} &30.71 &0.903 &33.17& \underline{0.951} & 34.57 &0.978 & 29.63 &0.882& 31.39 & 0.885 & 31.89 & 0.920
        \\
        ${\color{blue}\blacklozenge}$BaryIR~\cite{baryir11417902} & 30.82 &\textbf{0.907} & 33.05& 0.948& \underline{34.60}& 0.979 & 29.52& 0.881& 31.32 & 0.884 & 31.86 & 0.920
        \\
           ${\color{blue}\blacklozenge}$Perceive-IR~\cite{Perceive-IR10990319}& \underline{30.93} & 0.902 & 32.99 & 0.948 & 34.52 & \underline{0.980} & \underline{29.79} & \underline{0.889} & 31.44 & 0.885 & \underline{31.93} & \underline{0.921}
        \\
      \rowcolor{gray!20} \textbf{Perceive-IR$^G$}~\cite{Perceive-IR10990319}&\textbf{31.08} &\textbf{0.907} & \textbf{33.24}& \textbf{0.952}& \textbf{34.79}& \textbf{0.981} & \textbf{29.96}& \textbf{0.891}& \textbf{31.63} & \textbf{0.887} & \textbf{32.14} & \textbf{0.924}
      \\
    \hline
\end{tabular}}
\end{table*}

\begin{figure*}
    \centering
    \includegraphics[width=\linewidth]{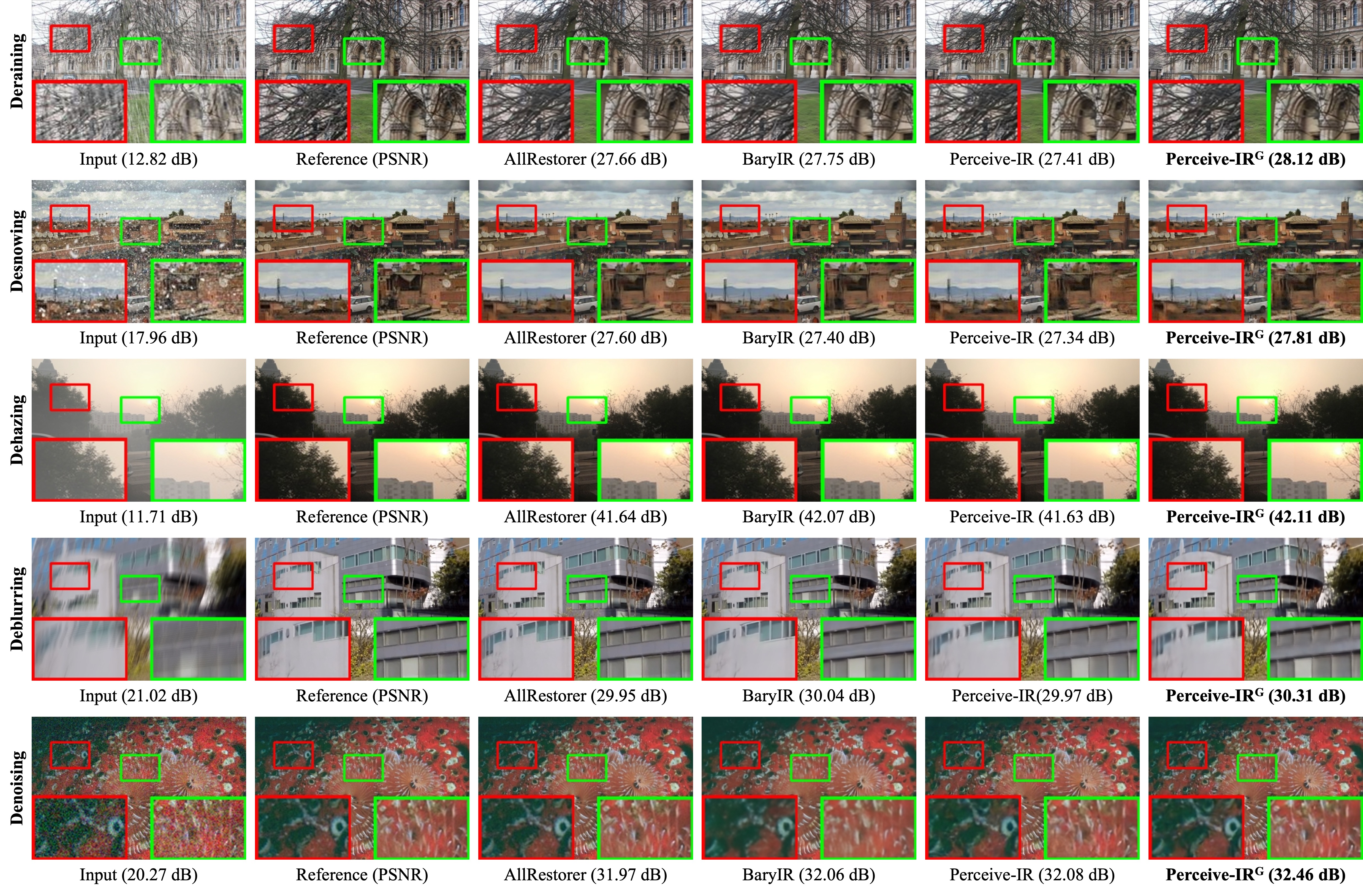}
    \caption{
    Qualitative comparison in the all-in-one image restoration setting.
    }
    \label{fig:qualitative_allinone}
\end{figure*}

\subsubsection{Capability Expansion toward All-in-One Restoration}
\label{sec:allinone_results}

We further investigate whether GraLoD can extend existing restoration backbones toward all-in-one restoration under multi-degradation training. Table~\ref{tb:allinone} compares task-specific, task-aligned, and dedicated all-in-one methods under the same five-degradation setting. 
The most significant gain is obtained with ALGNet~\cite{ALGgao2024learning}. After introducing GraLoD, its average performance improves by 1.10 dB, with improvements observed across all five restoration tasks. Particularly large gains are obtained for deraining, desnowing, and dehazing, reaching approximately 1.3--2.0 dB. This result indicates that a task-specific backbone can be effectively adapted to heterogeneous restoration requirements through spatially and stage-wise scale selection, without introducing degradation-specific branches or redesigning the backbone into a dedicated all-in-one architecture.

A similar trend is observed for the task-aligned StarIR~\cite{starir11429607} backbone. GraLoD improves its average performance by $0.38$ dB, with particularly evident improvement on deblurring. More importantly, StarIR$^{G}$ becomes competitive with dedicated all-in-one approaches and surpasses several of them in average restoration performance. 
GraLoD also remains beneficial when applied to a model already designed for all-in-one restoration. Adding GraLoD to Perceive-IR~\cite{Perceive-IR10990319} further improves its average performance by $0.21$ dB and yields consistent gains across all five degradation types, resulting in the best overall performance among the compared methods. 
Figure~\ref{fig:qualitative_allinone} presents qualitative comparisons under the all-in-one restoration setting. Across all five degradation types, integrating GraLoD into Perceive-IR~\cite{Perceive-IR10990319} consistently improves visual restoration quality over the original model and remains competitive with or superior to dedicated all-in-one methods.

\subsubsection{Zero-Shot Generalization to Real-World Degradations}
\label{sec:zero_shot}

We  evaluate the zero-shot generalization ability of GraLoD on real-world degradation datasets that are not used during training. As shown in Table~\ref{tb:allrealw}, integrating GraLoD into Perceive-IR~\cite{Perceive-IR10990319} consistently improves its performance across real-world rain, haze, and noise degradations.
On RealRain-1k-L~\cite{realrain1li2022toward}, Perceive-IR$^{G}$ improves PSNR by $0.30$ dB while also achieving better SSIM and LPIPS, indicating that GraLoD improves both reconstruction fidelity and perceptual quality under real rain patterns. Similar improvements are observed on RTTS~\cite{Rttsli2018benchmarking}, where GraLoD consistently improves all three no-reference quality measures. These results suggest that the learned LOD adaptation remains effective even when the spatial characteristics of real-world haze differ from those encountered during training. On SIDD~\cite{ssidabdelhamed2018high}, GraLoD further improves PSNR while maintaining the best SSIM, demonstrating that its benefit also transfers to real sensor noise.

\begin{table*}
\centering
\caption{Zero-shot generalization results on real-world degradation datasets.}
\label{tb:allrealw}
\resizebox{\linewidth}{!}{
\begin{tabular}{c|ccc|ccc|cc}
    \hline
    \multirow{2}{*}{Methods} & \multicolumn{3}{c|}{RealRain-1k-L} & \multicolumn{3}{c|}{RTTS} & \multicolumn{2}{c}{SIDD}
    \\
    &PSNR $\uparrow$ &  SSIM $\uparrow$ &LPIPS $\downarrow$ &FADE $\downarrow$ &BRISQUE $\downarrow$ &NIMA $\uparrow$&PSNR $\uparrow$ & SSIM $\uparrow$
    \\
    \hline\hline
   
  ECFNet~\cite{gao2026emphasizing} & 23.69 & 0.756 & 0.401 & 1.394 & 29.105 &4.372 & 24.33 & 0.471 
    \\
   
        Defusion~\cite{DefusionLuo_2025_CVPR}& \underline{27.37} & \underline{0.903} & \underline{0.371} & 1.277 & \underline{23.520} & \underline{4.615} & 24.52 & \underline{0.495}
        \\
         Perceive-IR~\cite{Perceive-IR10990319}& 27.31 & 0.901  & 0.372 & \underline{1.264} & 23.694 &4.613 & \underline{24.53} & \textbf{0.497}
        \\
      \rowcolor{gray!20} \textbf{Perceive-IR$^G$}~\cite{Perceive-IR10990319} & \textbf{27.61} & \textbf{0.904}  & \textbf{0.366} & \textbf{1.243} & \textbf{23.489} &\textbf{4.620} & \textbf{24.59} & \textbf{0.497}
    \\
    \hline
\end{tabular}}
\end{table*}

\subsection{Ablation Studies}
\subsubsection{Effect of Each Component}

\begin{table}
    \centering

\caption{Ablation study on the individual components of GraLoD.}

    \label{tab:abl1}
    \begin{tabular}{ccccccc}
    \hline
        Settings& LOD & LFA & MSFC & SAR &  PSNR &$\triangle$ PSNR 
         \\
         \hline
         \hline
         (a)  & & & & & 30.60 & -
         \\
         (b)  &\ding{52}& &&  & 31.01 & + 0.41 dB
         \\
          (c)  &\ding{52} &\ding{52} &&  & 31.28 & + 0.68 dB
         \\
        (d)  &\ding{52} &\ding{52} &\ding{52}&  & 31.49 & + 0.89 dB
         \\
          (e)  &\ding{52} &\ding{52} &\ding{52} &\ding{52} & 31.70 & + 1.10 dB
        \\
         \hline
    \end{tabular}
\end{table}

We investigate the contribution of the main components of GraLoD using ALGNet~\cite{ALGgao2024learning} as the base restoration model. As shown in Table~\ref{tab:abl1}, all components progressively improve the restoration performance over the original backbone. For the LOD-only variant, a minimal channel projection is used to make the native multi-scale encoder features dimensionally compatible, while the complete LOD feature alignment (LFA) proposed in GraLoD is disabled.

Introducing LOD  alone improves the baseline by $0.41$ dB, demonstrating that explicitly adapting the representation scale during decoding is beneficial even without the complete GraLoD design. This result supports our central motivation that the spatial support required for restoration should not be determined solely by the fixed multi-scale hierarchy of the backbone.
Adding LFA further improves the performance by $0.27$ dB, resulting in a cumulative gain of $0.68$ dB over the baseline. This indicates that continuous scale querying benefits from first mapping heterogeneous encoder features into a better aligned LOD representation space. 
MSFC provides an additional $0.21$ dB improvement. Unlike reconstruction supervision alone, which does not uniquely constrain the predicted scale assignment, MSFC encourages each spatial location to select the smallest representation footprint that provides sufficient restoration utility. The improvement suggests that explicitly calibrating the learned LOD field helps prevent arbitrary or biased scale selection.
Finally, incorporating SAR yields a further $0.21$ dB gain, bringing the overall improvement to $1.10$ dB over the original backbone. This confirms that considering the spatial organization of the LOD field is complementary to scale calibration. MSFC determines which representation scale is locally appropriate, whereas SAR encourages these assignments to remain coherent within homogeneous regions while allowing transitions around structural boundaries.

\subsubsection{Continuous LOD versus Conventional Multi-Scale Adaptation}
\label{sec:ablation_scale}

A central distinction between GraLoD and conventional multi-scale restoration lies in how representations at different scales are utilized. 
As shown in Table~\ref{tb:ablation_scale}, relying exclusively on either the finest or coarsest representation provides only limited improvements over the original backbone. Uniformly averaging the multi-scale features also yields a relatively small gain, indicating that simply increasing access to features from different resolutions is insufficient. Replacing fixed aggregation with learnable concatenation and convolution further improves the performance, demonstrating the benefit of adaptive multi-scale feature integration.
A larger improvement is obtained with scale attention, which outperforms fixed-scale and conventional fusion strategies by dynamically adjusting the contribution of different feature levels. This observation confirms that the appropriate representation scale varies with image content and cannot be adequately captured by a fixed aggregation scheme.
GraLoD further improves upon scale attention by $0.48$ dB. Unlike attention-based fusion, which independently assigns weights to different feature levels, GraLoD preserves the ordering of the scale hierarchy and performs scale selection within a continuously queryable LOD space.

\begin{table}
\centering
\caption{Comparison of different multi-scale adaptation strategies.}
\label{tb:ablation_scale}
\resizebox{\linewidth}{!}{
\begin{tabular}{c|cccccc}
\hline
Scale Adaptation Strategy & Finest only & Coarsest only &Average & Concat + Conv & Attention & GraLoD
\\
\hline\hline
PSNR& 30.83 &30.71 &30.76 &30.94 &31.22  &31.70
\\
\hline
\end{tabular}}
\end{table}

\section{Conclusion}
\label{sec:conclusion}

In this work, we introduced GraLoD, a plug-and-play framework that treats restoration scale as a spatially varying and stage-dependent continuous variable. GraLoD reuses the native encoder hierarchy, aligns it into an ordered LOD representation space, and continuously queries the representation scale required at each decoder stage. Minimal-Sufficient Footprint Calibration and structure-aware regularization further constrain the learned LOD field to remain restoration-effective and spatially meaningful.
Extensive experiments demonstrate that GraLoD consistently improves existing task-specific restoration models and, under multi-degradation training, enables conventional backbones to be extended toward competitive all-in-one restoration. 

\section*{AI Use Statement}

In this work, we used generative AI tools to assist with manuscript editing. GPT-based tools were used to improve the language, grammar, and readability of the manuscript without changing the underlying scientific content.  The underlying data and numerical results in the figure were not generated or modified by AI.

All AI-assisted outputs were manually reviewed and verified by the authors.
The authors take full responsibility for the final manuscript, including all
text, figures, experimental results, scientific claims, and conclusions.

\bibliography{iclr2027_conference}
\bibliographystyle{iclr2027_conference}

\appendix
\section{Appendix}

\subsection{Additional Method Details}
\label{app:method}

\subsubsection{Properties of Continuous LOD Querying}
\label{app:continuous_lod}

As described in Sec.~\ref{sec:gralod}, GraLoD represents the required
representation scale at decoder stage $s$ by a continuous LOD coordinate
$\lambda_s(\mathbf{p})\in[0,L-1]$. Unlike multi-scale attention that assigns
independent weights to multiple feature levels, this coordinate lies on an
ordered scale axis and determines only two neighboring levels for
interpolation. We further discuss the continuity, differentiability, and
stage-conditioned nature of this formulation below.

\paragraph{Continuous interpolation along the LOD axis.}
Consider a spatial location $\mathbf{p}$ for which
$\lambda_s(\mathbf{p})\in[l,l+1]$, where
$l\in\{0,\ldots,L-2\}$. According to Eq.~\ref{eq:continuous_lod}, the two
neighboring levels are $l$ and $l+1$, and the queried representation can be
written as
\begin{equation}
\begin{split}
F_{\mathrm{LOD}}^s(\mathbf{p})
={}&
\big(l+1-\lambda_s(\mathbf{p})\big)Z_s^l(\mathbf{p})
\\
&+
\big(\lambda_s(\mathbf{p})-l\big)Z_s^{l+1}(\mathbf{p}).
\end{split}
\label{eq:app_lod_interp}
\end{equation}
The interpolation coefficients are non-negative and sum to one. Therefore,
$F_{\mathrm{LOD}}^s(\mathbf{p})$ remains a convex combination of two
neighboring representations.

The same operation can also be expressed using a triangular basis:
\begin{equation}
F_{\mathrm{LOD}}^s(\mathbf{p})
=
\sum_{l=0}^{L-1}
\omega_l\big(\lambda_s(\mathbf{p})\big)
Z_s^l(\mathbf{p}),
\label{eq:app_triangular_query}
\end{equation}
where $\omega_l(\lambda) = \max\big(0,1-|\lambda-l|\big)$. For any $\lambda\in[0,L-1]$, $\sum_{l=0}^{L-1}\omega_l(\lambda)=1$, and at most two adjacent weights are non-zero. This differs from generic
scale attention, where all feature levels may receive independently learned
weights. GraLoD instead restricts the query to a local neighborhood along the
ordered scale axis.

\paragraph{Continuity.}
The queried representation changes continuously with the LOD coordinate.
For an integer LOD value $k$, the limits from the two adjacent intervals
satisfy
\begin{equation}
\lim_{\lambda\rightarrow k^-}
F_{\mathrm{LOD}}^s(\mathbf{p})
=
Z_s^k(\mathbf{p})
=
\lim_{\lambda\rightarrow k^+}
F_{\mathrm{LOD}}^s(\mathbf{p}).
\label{eq:app_lod_continuity}
\end{equation}
Thus, moving the predicted LOD coordinate across an integer boundary does
not introduce a discontinuous change in the queried representation. This
property avoids the abrupt feature switching associated with hard
discrete-scale selection.

\paragraph{Gradient with respect to the LOD coordinate.}
For a non-integer coordinate
$\lambda_s(\mathbf{p})\in(l,l+1)$, differentiating
Eq.~\ref{eq:app_lod_interp} gives
\begin{equation}
\frac{
\partial F_{\mathrm{LOD}}^s(\mathbf{p})
}{
\partial \lambda_s(\mathbf{p})
}
=
Z_s^{l+1}(\mathbf{p})
-
Z_s^l(\mathbf{p}).
\label{eq:app_lod_gradient}
\end{equation}
The reconstruction objective can therefore propagate gradients through the
continuous query to the LOD estimator $\mathcal{G}_s$. The derivative is
piecewise defined because the neighboring levels change at integer
coordinates, while the queried representation itself remains continuous as
shown above.

\paragraph{Why stage-conditioned LOD prediction?}
The representation scale required at a spatial location can change as
reconstruction proceeds. A decoder feature $D_s$ contains the intermediate
reconstruction state available at stage $s$, and conditioning the estimator
on $D_s$ therefore allows GraLoD to update the LOD requirement at different
stages:
\begin{equation}
\boldsymbol{\lambda}_s
=\mathcal{G}_s(D_s,Z_s^0,\ldots,Z_s^{L-1}).
\end{equation}
A single LOD field shared by all decoder stages would instead assume that the
required representation scale remains unchanged throughout reconstruction.
Stage-conditioned prediction removes this restriction and allows each stage
to determine its own scale requirements from the current feature state.
This design does not impose a monotonic coarse-to-fine constraint on
$\boldsymbol{\lambda}_s$. Although broader contextual representations may be
useful for recovering large structures at some stages and finer
representations may become more useful for local refinement at others, the
LOD coordinate is learned independently at each decoder stage according to
the image content and reconstruction state.

\subsubsection{Integration into Existing Restoration Backbones}
\label{app:integration}

GraLoD is designed as an auxiliary representation-scale adaptation mechanism
rather than a new restoration backbone. It reuses the hierarchical features
already produced by the encoder and leaves the original feature-processing
blocks and skip connections unchanged. This section describes the feature
alignment, resolution alignment, and decoder integration used in our
implementation.

\paragraph{Feature alignment.}
Let $F^l\in \mathbb{R}^{H_l\times W_l\times C_l}$, $l=0,\ldots,L-1$, denote the native encoder features. Since their channel dimensions and
representation characteristics differ across levels, each feature is first
mapped into a common $C$-channel space as defined in
Eq.~\ref{eq:lod_alignment}:
\begin{equation}
Z^l=\mathcal{A}_l(F^l).
\end{equation}
A lightweight implementation of $\mathcal{A}_l$ is
\begin{equation}
\mathcal{A}_l
=
LN \circ \operatorname{Conv}_{1\times1}
\circ
\operatorname{DWConv}_{3\times3}
\circ
SG \circ SCA \circ
\operatorname{Conv}_{1\times1},
\label{eq:app_feature_alignment}
\end{equation}
where the first projection aligns the channel dimension, the depth-wise
convolution, SG~\cite{chen2022simple} and SCA~\cite{chen2022simple} performs local refinement, and the final projection produces the
aligned feature representation. Other lightweight projection blocks can be
used without changing the GraLoD formulation.
The resulting hierarchy $\mathcal{Z} = \{Z^0,Z^1,\ldots,Z^{L-1}\} $ is constructed once and shared by all decoder stages. Importantly, GraLoD
does not construct an additional image or feature pyramid, it organizes the
native encoder hierarchy into an ordered and queryable LOD representation
space.

\paragraph{Resolution alignment.}
The aligned features $Z^l$ retain the spatial resolutions of their
corresponding encoder levels. Before LOD estimation and interpolation at
decoder stage $s$, they are mapped to the spatial lattice
$\Omega_s$ according to Eq.~\ref{eq:stage_alignment}. We implement
$\mathcal{R}_{s,l}$ as
\begin{equation}
\mathcal{R}_{s,l}(Z^l)
=
\begin{cases}
\operatorname{Up}_{s,l}(Z^l),
&
H_l < H_s,
\\[3pt]
Z^l,
&
H_l = H_s,
\\[3pt]
\operatorname{Down}_{s,l}
\big(
K_{\mathrm{aa}} * Z^l
\big),
&
H_l > H_s,
\end{cases}
\label{eq:app_resolution_alignment}
\end{equation}
where $H_s\times W_s$ denotes the spatial resolution of decoder stage $s$,
$K_{\mathrm{aa}}$ is a low-pass anti-aliasing kernel, and
$\operatorname{Up}_{s,l}$ and $\operatorname{Down}_{s,l}$ denote the
corresponding resizing operations. Anti-aliased filtering is applied before
spatial reduction to suppress aliasing across LOD levels. Standard
interpolation is sufficient when an aligned feature is upsampled to the
decoder resolution.

After this operation, $Z_s^l = \mathcal{R}_{s,l}(Z^l) \in \mathbb{R}^{H_s\times W_s\times C}$, so that all LOD levels are spatially compatible with $D_s$ and can be used jointly by the LOD estimator.

\paragraph{Stage-wise LOD estimation and query.}
At decoder stage $s$, the current decoder feature and aligned LOD features
are concatenated to predict
$\boldsymbol{\lambda}_s$. For every location $\mathbf{p}$, the predicted
coordinate determines the neighboring levels
$l_s^-(\mathbf{p})$ and $l_s^+(\mathbf{p})$, which are interpolated according
to Eq.~\ref{eq:continuous_lod}. Only two neighboring levels contribute to
$F_{\mathrm{LOD}}^s(\mathbf{p})$, regardless of the total number of levels
$L$.

The aligned hierarchy $\mathcal{Z}$ is shared globally across decoder stages,
while $\mathcal{G}_s$ is stage-specific. This separation avoids repeatedly
constructing scale representations while allowing the scale query to adapt
to the reconstruction state at each stage.

\paragraph{Residual decoder integration.}
The queried feature is incorporated through the residual adaptor defined in
Eq.~\ref{eq:stage_adapter}:
\begin{equation}
\widetilde{D}_s
= D_s + \mathcal{H}_s(D_s,F_{\mathrm{LOD}}^s),
\label{eq:stage_adapter}
\end{equation}
The residual form allows GraLoD to augment the original decoder
representation without replacing it. The enhanced feature
$\widetilde{D}_s$ is subsequently processed by the native decoder block
together with the original skip feature. Consequently, the backbone-specific
encoder blocks, decoder blocks, and skip topology remain unchanged.

\subsubsection{Further Analysis of Minimal-Sufficient Footprint Calibration}
\label{app:msfc}

As introduced in Sec.~\ref{sec:lod_regularization}, reconstruction supervision
alone does not uniquely determine the LOD coordinate. MSFC provides an
auxiliary target by comparing the local restoration utility of the aligned LOD
levels while penalizing unnecessarily coarse representations.

\paragraph{Scale probes.}
At decoder stage $s$, each aligned representation $Z_s^l$ is evaluated using
a lightweight probe shared across all LOD levels:
\begin{equation}
\hat{\mathbf{x}}_s^{\,l}
=
\mathcal{P}_s
\left(
\operatorname{Concat}
\left(
D_s,Z_s^l
\right)
\right),
\qquad
l=0,\ldots,L-1.
\label{eq:app_probe}
\end{equation}
Sharing $\mathcal{P}_s$ across levels prevents the comparison from being
affected by level-specific prediction heads. The probes are auxiliary modules
used only to estimate the relative restoration utility of different
representation scales.
The local reconstruction error is computed as:
\begin{equation}
e_{s,l}(\mathbf{p})
=
\mathcal{A}_r
\left(
\left|
\hat{\mathbf{x}}_s^{\,l}
-
\mathbf{x}_s
\right|
\right)(\mathbf{p}),
\label{eq:app_local_error}
\end{equation}
where $\mathcal{A}_r$ averages the reconstruction error over a local
neighborhood centered at $\mathbf{p}$. Local averaging makes the estimated
scale requirement depend on a spatial region rather than an isolated pixel.

\paragraph{Minimal-sufficient criterion.}
The scale cost combines the reconstruction error with a penalty that increases
toward coarser LOD levels:
\begin{equation}
c_{s,l}(\mathbf{p})
=
e_{s,l}(\mathbf{p})
+
\gamma\frac{l}{L-1}.
\label{eq:app_scale_cost}
\end{equation}
Consider two levels $l$ and $k$ with $k>l$. The coarser level $k$ is preferred
over $l$ only if
\begin{equation}
c_{s,k}(\mathbf{p})
<
c_{s,l}(\mathbf{p}),
\end{equation}
which is equivalent to
\begin{equation}
e_{s,l}(\mathbf{p})
-
e_{s,k}(\mathbf{p})
>
\gamma\frac{k-l}{L-1}.
\label{eq:app_minimal_sufficient}
\end{equation}
Therefore, moving to a coarser representation is beneficial only when the
reduction in local reconstruction error is large enough to compensate for the
additional scale cost. This gives the ``minimal-sufficient'' interpretation of
MSFC: broader spatial support is selected only when it provides sufficient
restoration benefit.

\paragraph{Soft LOD target.}
Instead of selecting the minimum-cost level through a hard discrete operation,
we convert the scale costs into a soft distribution:
\begin{equation}
q_{s,l}(\mathbf{p})
=
\frac{
\exp\left(-c_{s,l}(\mathbf{p})/\tau\right)
}{
\sum_{j=0}^{L-1}
\exp\left(-c_{s,j}(\mathbf{p})/\tau\right)
}.
\label{eq:app_scale_distribution}
\end{equation}
The corresponding continuous target coordinate is:
\begin{equation}
\lambda_s^*(\mathbf{p})
=
\sum_{l=0}^{L-1}
l\,q_{s,l}(\mathbf{p}).
\label{eq:app_target_lod}
\end{equation}
The temperature $\tau$ controls the concentration of the distribution. A
smaller $\tau$ produces a target closer to discrete scale selection, while a
larger value distributes probability over a wider range of neighboring
levels. In the limit $\tau\rightarrow0$, the soft target approaches the
minimum-cost LOD level when the minimum is unique.

The stop-gradient operation in Eq.~\ref{eq:calibration} prevents the predicted
LOD field from changing its own calibration target through the target
construction. The target therefore serves as auxiliary supervision for the
LOD estimator rather than forming a trivial self-consistent solution.

\paragraph{Probe training.}
The probes are trained using the resized clean target:
\begin{equation}
\mathcal{L}_{\mathrm{probe}}
=
\frac{1}{S}
\sum_{s=1}^{S}
\frac{1}{L}
\sum_{l=0}^{L-1}
\frac{1}{|\Omega_s|}
\left\|
\hat{\mathbf{x}}_s^{\,l}
-
\mathbf{x}_s
\right\|_1.
\label{eq:probe_loss}
\end{equation}
To keep the probes auxiliary, we detach their inputs from the restoration
features during probe optimization:
\begin{equation}
\hat{\mathbf{x}}_s^{\,l}
=
\mathcal{P}_s
\left(
\operatorname{sg}
\left[
\operatorname{Concat}
(D_s,Z_s^l)
\right]
\right).
\label{eq:app_detached_probe}
\end{equation}
In this case, $\mathcal{L}_{\mathrm{probe}}$ updates the probe parameters
without encouraging the restoration backbone to modify its representations
solely to simplify the auxiliary prediction task. All probes are discarded
after training.

\subsubsection{Analysis of Structure-Aware LOD Regularization}
\label{app:structure}

MSFC determines which representation scale is locally useful, but applying the
calibration independently at each location does not explicitly constrain the
spatial organization of the predicted LOD field. Neighboring locations
belonging to the same image structure are generally expected to require
similar spatial support, whereas scale transitions should remain possible near
structural boundaries.

For neighboring locations $\mathbf{p}$ and $\mathbf{q}$ at decoder stage $s$,
we define the structure affinity as
\begin{equation}
w_{\mathbf{p}\mathbf{q}}^s
=
\exp
\left(
-\beta
\left\|
\mathbf{x}_s(\mathbf{p})
-
\mathbf{x}_s(\mathbf{q})
\right\|_1
\right).
\label{eq:app_structure_weight}
\end{equation}
When two neighboring locations have similar clean-image content,
$w_{\mathbf{p}\mathbf{q}}^s$ approaches one, imposing a stronger penalty on
differences between their LOD coordinates. Across an image boundary, the
appearance difference increases and the corresponding affinity decreases,
allowing the LOD field to vary more freely.
The regularization term in Eq.~\ref{eq:structure_reg} can therefore be viewed
as an edge-aware total-variation constraint on the spatial LOD field:
\begin{equation}
\mathcal{L}_{\mathrm{str}}
=
\frac{1}{S}
\sum_{s=1}^{S}
\frac{1}{|\Omega_s|}
\sum_{\mathbf{p}\in\Omega_s}
\sum_{\mathbf{q}\in\mathcal{N}(\mathbf{p})}
w_{\mathbf{p}\mathbf{q}}^s
\left|
\lambda_s(\mathbf{p})
-
\lambda_s(\mathbf{q})
\right|.
\end{equation}
Unlike uniform smoothness regularization, the structure-aware weighting does
not force the LOD field to be globally smooth. It favors locally coherent
representation scales while preserving transitions around image structures.
We use the resized clean target $\mathbf{x}_s$ to construct the affinity only
during training. Using the degraded observation directly could introduce
degradation-induced edges into the regularization weights, causing rain
streaks, noise, or other corruptions to be interpreted as scene boundaries.
No clean image or structure weighting is required during inference.

\subsubsection{Training Strategy}
\label{app:training}

GraLoD is trained jointly with the underlying restoration backbone using
Eq.~\ref{eq:total_objective}. The reconstruction term
$\mathcal{L}_{\mathrm{rec}}$ follows the original objective of each backbone,
while $\mathcal{L}_{\mathrm{cal}}$ and $\mathcal{L}_{\mathrm{str}}$ supervise
the predicted LOD fields. The auxiliary probes are optimized with
$\mathcal{L}_{\mathrm{probe}}$ and are removed after training.

At the beginning of training, the scale probes have not yet learned reliable
estimates of the restoration utility of different LOD levels. We therefore
allow the probe predictions to stabilize before applying strong calibration.
The calibration coefficient can be gradually increased as
\begin{equation}
\lambda_{\mathrm{cal}}(t)
=
\lambda_{\mathrm{cal}}^{\max}r(t),
\label{eq:app_calibration_schedule}
\end{equation}
where $t$ denotes the training iteration and $r(t)$ increases from $0$ to $1$
during the warm-up period.
For each decoder stage, the clean image is resized to the corresponding
resolution to obtain $\mathbf{x}_s$. Anti-aliased downsampling is used when
spatial reduction is required. The same training procedure is used for
task-specific and multi-degradation settings, without introducing
degradation labels or task-dependent GraLoD branches.
At inference, the clean targets, scale probes, MSFC target construction, and
structure-aware weights are all removed. Only the shared LOD alignment,
stage-specific LOD estimators, continuous queries, and residual adaptors are
retained. 

\subsubsection{Computational Complexity}
\label{app:complexity}

GraLoD reuses the multi-scale hierarchy already produced by the restoration
backbone and therefore does not construct an additional feature pyramid. Its
inference cost can be written as
\begin{equation}
\mathcal{C}_{\mathrm{GraLoD}}
=
\mathcal{C}_{\mathrm{base}}
+
\mathcal{C}_{\mathrm{align}}
+
\sum_{s=1}^{S}
\left(
\mathcal{C}_{\mathrm{LOD}}^{s}
+
\mathcal{C}_{\mathrm{adapt}}^{s}
\right),
\label{eq:app_complexity}
\end{equation}
where $\mathcal{C}_{\mathrm{align}}$ denotes the cost of constructing the
shared aligned hierarchy, $\mathcal{C}_{\mathrm{LOD}}^{s}$ is the cost of
stage-wise LOD estimation and querying, and
$\mathcal{C}_{\mathrm{adapt}}^{s}$ denotes the residual adaptor overhead.
The aligned hierarchy is constructed once and reused across decoder stages.
The auxiliary probes and all target-dependent computations in MSFC and
structure-aware regularization are required only during training and therefore
do not contribute to inference-time complexity.

\subsection{Experimental Setup}
\label{sec:expsetup}

\subsubsection{Datasets}

\textbf{i) Image deraining.}
For image deraining, we use 13,712 paired clean--rain images collected from multiple datasets~\cite{Rain100,Test100,8099669,7780668} for training.
Evaluation is conducted on four commonly used benchmarks: Rain100H~\cite{Rain100}, Rain100L~\cite{Rain100}, Test100~\cite{Test100}, and Test1200~\cite{DIDMDN}.

\textbf{ii) Image desnowing.}
For image desnowing, we use Snow100K~\cite{desnownet}, SRRS~\cite{JSTASRchen2020jstasr}, and CSD~\cite{HDCW-Netchen2021all}.
Following the protocol of previous work~\cite{FSNet}, 2,500 paired images are randomly sampled for training and 2,000 images are used for evaluation.

\textbf{iii) Image dehazing.}
For image dehazing, we use the daytime synthetic subsets of RESIDE~\cite{RESIDEli2018benchmarking}, including the Indoor Training Set (ITS), Outdoor Training Set (OTS), and Synthetic Objective Testing Set (SOTS).
Models are trained separately on ITS and OTS and evaluated on SOTS-Indoor and SOTS-Outdoor, respectively, each containing 500 paired images.

\textbf{iv) Image deblurring.}
For image deblurring, we use the GoPro dataset~\cite{Gopro}, which contains 2,103 training pairs and 1,111 testing pairs.

\textbf{v) Image denoising.}
For image denoising, we construct a composite training set using 800 images from DIV2K~\cite{DIK}, 2,650 images from Flickr2K~\cite{lim2017enhanced}, 400 images from BSD500~\cite{BSD500}, and 4,744 images from WED~\cite{ma2016waterloo}.
Additive white Gaussian noise is synthesized with the noise level $\sigma$ randomly sampled from $\{15,25,50\}$.
Evaluation is conducted on CBSD68~\cite{BSD68}, Urban100~\cite{urban100}, and Kodak24~\cite{kodak}.

For the all-in-one setting, the training sets of the above restoration tasks are combined to form a mixed training set.
For unified quantitative comparison, we report results on Test100~\cite{Test100} for deraining, Snow100K~\cite{desnownet} for desnowing, SOTS-Outdoor~\cite{RESIDEli2018benchmarking} for dehazing, GoPro~\cite{Gopro} for deblurring, and CBSD68~\cite{BSD68} with $\sigma=25$ for denoising.

\subsubsection{Default Hyperparameters}
\label{app:hyperparameters}

Unless otherwise specified, GraLoD is applied to all decoder stages of the
restoration backbone. The native encoder hierarchy is reused to construct the
shared LOD representation space, and the number of LOD levels $L$ follows the
number of selected encoder stages. For backbones with four major encoder
stages, we use $L=4$. All aligned features are projected to the deepest encoder feature channel
dimension through $\mathcal{A}_l$.

For MSFC, the scale penalty is set to
$\gamma=0.05$ and the temperature of the soft LOD distribution is set to
$\tau=0.20$. The local restoration error is computed using a $5\times5$
averaging window. For structure-aware regularization, we set $\beta=10$.
The loss weights are fixed as
$\lambda_{\mathrm{cal}}=0.10$,
$\lambda_{\mathrm{str}}=0.05$, and
$\lambda_{\mathrm{probe}}=0.10$.
The reconstruction loss retains a weight of $1$.

The same GraLoD configuration is used across CNN-, Transformer-, and
state-space-based restoration backbones unless otherwise specified.
The aligned LOD hierarchy is shared across decoder stages, whereas
$\mathcal{G}_s$ and $\mathcal{H}_s$ are instantiated independently for each
stage. All scale probes and target-dependent computations used by MSFC and
structure-aware regularization are removed at inference.

\subsubsection{Evaluation Metrics} We evaluate performance using both reference-based and no-reference metrics. The reference-based metrics include Peak Signal-to-Noise Ratio (PSNR), Structural Similarity Index (SSIM), and Learned Perceptual Image Patch Similarity (LPIPS)~\cite{54zhang2018unreasonable}. The no-reference metrics include the Underwater Colour Image Quality Evaluation Metric (UCIQE)~\cite{55yang2015underwater}, Underwater Image Quality Measure (UIQM)~\cite{56panetta2015human}, Fog Aware Density Evaluator (FADE)~\cite{58choi2015referenceless}, Blind/Referenceless Image Spatial Quality Evaluator (BRISQUE)~\cite{59mittal2011blind}, and Neural Image Assessment (NIMA)~\cite{60talebi2018nima}. Among them, UCIQE and UIQM are specifically designed for underwater image restoration evaluation, while FADE, BRISQUE, and NIMA are commonly used to assess dehazing performance in real-world scenarios. For PSNR, SSIM, UCIQE, UIQM, and NIMA, higher values indicate better performance, whereas lower values are preferred for LPIPS, FADE, and BRISQUE. In the tables, the best and second-best results are highlighted in bold and underlined, respectively.

\subsection{More Experiments}
\label{sec:expset}

\begin{table}
\centering
\caption{Zero-shot generalization results on an unseen degradation type.}
\label{tb:allundt}
\begin{tabular}{c|ccc|cc}
    \hline
    \multirow{2}{*}{Methods} & \multicolumn{3}{c|}{UIEB} & \multicolumn{2}{c}{C60}
    \\
   & PSNR $\uparrow$ &  SSIM $\uparrow$ &LPIPS $\downarrow$ &UCIQE$\uparrow$ &UIQM$\uparrow$
    \\
    \hline\hline
ECFNet~\cite{gao2026emphasizing} & 20.49 & 0.856 & 0.223 & 0.539 & \underline{2.517} 
    \\
ACL~\cite{aclgu2025acl} & 20.94 & 0.867 & 0.201 & 0.552 & 2.485 
     \\

      Perceive-IR~\cite{Perceive-IR10990319}& \underline{21.77} &\underline{0.891} & \underline{0.169} & \underline{0.557} & \textbf{2.555}
        \\
      \rowcolor{gray!20}\textbf{Perceive-IR$^{G}$}~\cite{Perceive-IR10990319} & \textbf{21.98} & \textbf{0.893} & \textbf{0.162} & \textbf{0.559} & \textbf{2.555}
    \\
    \hline
\end{tabular}
\end{table}

\subsubsection{Generalization to an Unseen Degradation Type}
\label{sec:unseen_degradation} 
We further evaluate whether GraLoD can generalize beyond the degradation types observed during training by testing on underwater image enhancement datasets. As shown in Table~\ref{tb:allundt}, integrating GraLoD into Perceive-IR~\cite{Perceive-IR10990319} consistently improves its performance on the unseen underwater degradation domain. On UIEB~\cite{li2019underwater}, Perceive-IR$^{G}$ improves PSNR by $0.21$ dB while also achieving better SSIM and lower LPIPS, indicating improved reconstruction fidelity and perceptual quality despite the degradation type being absent from training. On C60~\cite{li2019underwater}, GraLoD further improves UCIQE while maintaining the best UIQM performance. These gains are obtained without introducing underwater-specific supervision, task identifiers, or additional adaptation modules. Compared with the baseline Perceive-IR and other competing methods, Perceive-IR$^{G}$ achieves the best overall performance across both benchmarks. This result suggests that the benefit of GraLoD is not limited to interpolation among degradation patterns seen during training. Instead, its continuous and stage-wise representation-scale adaptation enables the backbone to adjust its spatial support according to previously unseen image characteristics.

\begin{table}
\centering
\caption{Zero-shot generalization results on unseen noise levels.}
\label{tb:allunds}
\begin{tabular}{c|cccc}
    \hline
    \multirow{2}{*}{Methods} & \multicolumn{2}{c}{CBSD68} & \multicolumn{2}{c}{Urban100}
    \\
   & 60 & 100 & 60 & 100
    \\
    \hline\hline
     VLU-Net~\cite{VLUNetZeng_2025_CVPR}&  27.11 & 20.59 & 27.64 & 21.53
      \\
      Perceive-IR~\cite{Perceive-IR10990319}& \underline{27.13} & 20.65 & 27.65 & \underline{21.55}
        \\
        Defusion~\cite{DefusionLuo_2025_CVPR}& 27.11 & \underline{20.72} & \textbf{27.67} & 21.49
        \\
      \rowcolor{gray!20}\textbf{Perceive-IR$^G$}~\cite{Perceive-IR10990319} & \textbf{27.25} & \textbf{20.79} & \underline{27.66} & \textbf{21.63}
    \\
    \hline
\end{tabular}
\end{table}

\begin{figure*}
    \centering
    \includegraphics[width=\linewidth]{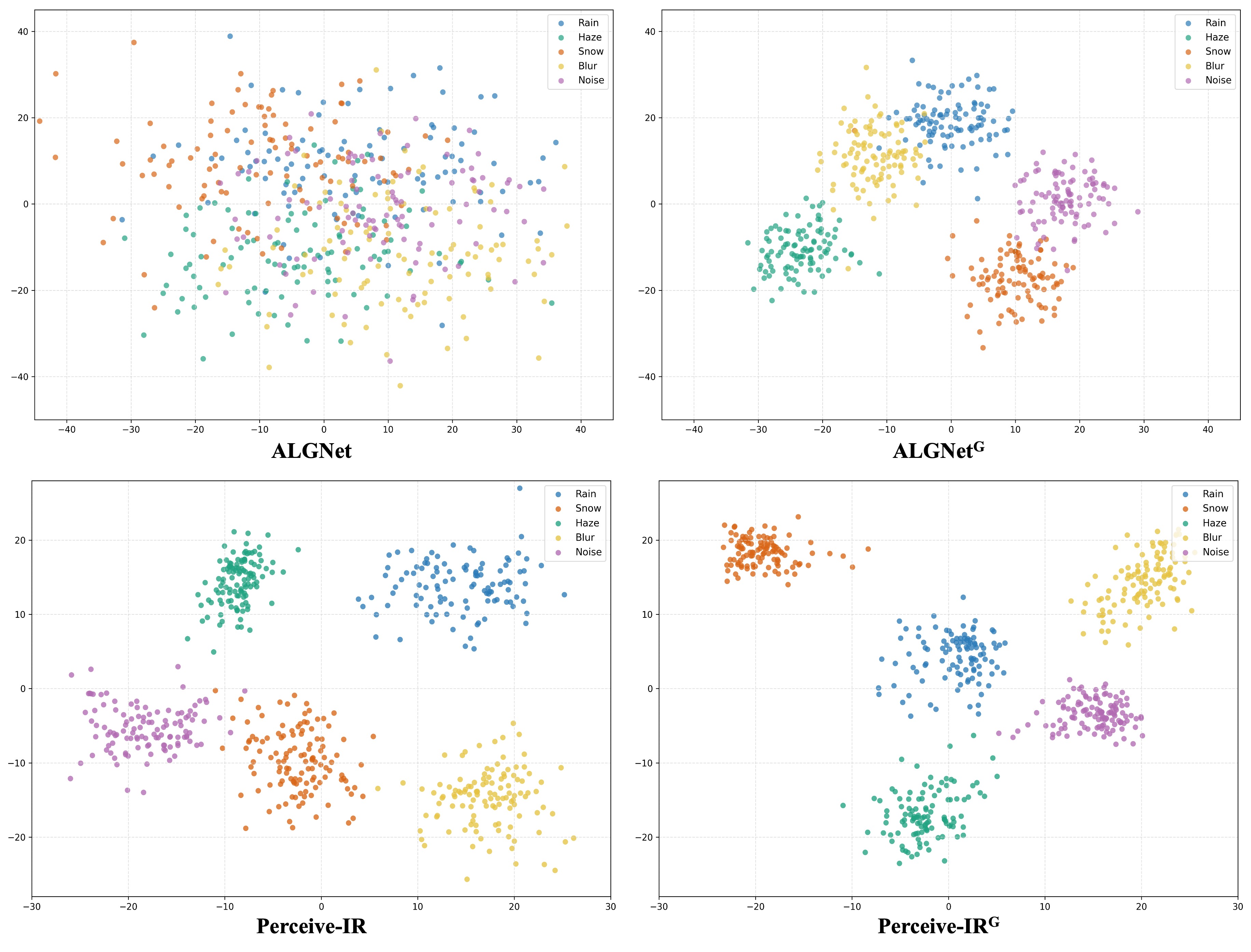}
    \caption{
    t-SNE visualization of feature representations for five degradation
    types with and without GraLoD.
    The top row compares ALGNet and ALGNet$^{G}$, while the bottom row compares
    Perceive-IR and Perceive-IR$^{G}$.
    GraLoD leads to substantially clearer degradation-dependent organization
    for the task-specific ALGNet backbone, whereas the global cluster structure
    of the all-in-one Perceive-IR model remains largely unchanged.
    }
    \label{fig:tsne}
\end{figure*}

\subsubsection{Generalization to Unseen Degradation Severities}
\label{sec:unseen_severity}

We further evaluate the robustness of GraLoD to unseen degradation severities by testing on noise levels that are not included during training. As shown in Table~\ref{tb:allunds}, integrating GraLoD into Perceive-IR~\cite{Perceive-IR10990319} consistently improves its denoising performance on both CBSD68~\cite{BSD68} and Urban100~\cite{urban100} under severe unseen noise levels.
On CBSD68, Perceive-IR$^{G}$ improves the baseline by $0.12$ dB at noise level 60 and by $0.14$ dB at noise level 100. Similar gains are observed on Urban100, where GraLoD improves the baseline under both unseen noise settings and achieves the best performance at the more challenging noise level 100. These improvements indicate that the learned LOD adaptation remains effective even when the degradation severity exceeds the range observed during training.
Compared with existing restoration methods, Perceive-IR$^{G}$ achieves the best or highly competitive performance across all evaluated settings. Notably, the advantage of GraLoD is maintained as the noise level increases, suggesting that its benefit is not tied to a fixed degradation intensity. Instead, the stage-wise LOD mechanism can adapt the effective spatial support according to the severity of the corrupted observation, allowing the backbone to access broader contextual information when stronger degradation requires more extensive spatial reasoning.

These results further demonstrate that GraLoD improves generalization not only across real-world degradation distributions and unseen degradation types, but also across previously unseen degradation severities.

\begin{table}
\centering
\caption{Effect of applying GraLoD to different decoder stages.}
\label{tb:ablation_stage_position}
\begin{tabular}{cccc|c}
\hline
$D_4$ & $D_3$ & $D_2$ & $D_1$ & PSNR $\uparrow$ \\
\hline\hline
\checkmark& & &  &  31.25\\
\checkmark &\checkmark&  & & 31.48 \\
\checkmark&\checkmark & \checkmark& & 31.61 \\
\checkmark&\checkmark & \checkmark & \checkmark & 31.70  \\
&\checkmark & \checkmark & & 31.23 \\
&\checkmark & \checkmark & \checkmark & 31.37  \\
\hline
\end{tabular}
\end{table}

\begin{table}
\centering
\caption{Comparison of different stage-wise LOD prediction strategies.}
\label{tb:ablation_stage_lod}
\begin{tabular}{l|c}
\hline
LOD Prediction & PSNR $\uparrow$ \\
\hline\hline
Stage-shared estimator & 30.96  \\
Stage-conditioned LOD & 31.70 \\
\hline
\end{tabular}
\end{table}

\subsubsection{Visualization of Degradation Representations}
\label{app:tsne}

To further examine how GraLoD affects the learned feature representations, we
visualize the feature distributions of different degradation types using
t-SNE. Figure~\ref{fig:tsne} compares the representations produced by ALGNet~\cite{ALGgao2024learning}
and Perceive-IR~\cite{Perceive-IR10990319} before and after integrating GraLoD. Five degradation types,
including rain, haze, snow, blur, and noise, are considered.

For the task-specific backbone ALGNet, the original representations show
substantial overlap among different degradation types, indicating that its
features are not explicitly organized according to degradation-dependent
characteristics. After integrating GraLoD, the five degradation categories
form substantially more compact and separable clusters. In particular, rain,
haze, snow, blur, and noise occupy distinct regions in the embedding space.
This observation suggests that continuous LOD adaptation helps a conventional
restoration backbone organize features according to the different spatial
support required by heterogeneous degradations. Such a change is consistent
with the strong improvement obtained when GraLoD is used to extend a
task-specific backbone toward multi-degradation restoration.

A different behavior is observed for Perceive-IR. Since Perceive-IR is
designed for all-in-one restoration, its original features already exhibit
clear degradation-dependent clustering. After GraLoD is introduced, the
overall cluster structure remains largely unchanged, although local feature
distributions are slightly reorganized. This indicates that the improvement
brought by GraLoD does not primarily arise from further separating degradation
categories. Instead, GraLoD complements the degradation-aware representation
already learned by the all-in-one model by improving how hierarchical
multi-scale features are selected and utilized during reconstruction.

These observations are consistent with the role of GraLoD in our formulation.
GraLoD does not introduce explicit degradation labels or an additional
degradation classification objective. It operates on the native encoder
hierarchy and adaptively queries the representation scale required at each
spatial location and decoder stage. Therefore, its effect can be more
pronounced for a backbone whose original feature space is not strongly
organized across degradation types, while for an all-in-one model with an
already discriminative degradation representation, the main benefit lies in
more effective representation-scale adaptation rather than further
restructuring the global degradation embedding.

\subsubsection{Effect of Stage-Wise LOD Adaptation}
\label{sec:ablation_stage}

GraLoD predicts an individual LOD field for each decoder stage rather than sharing one global scale assignment throughout reconstruction. We evaluate this design by applying GraLoD to different decoder stages and by comparing stage-conditioned LOD prediction with shared alternatives.
Table~\ref{tb:ablation_stage_position} evaluates GraLoD when it is inserted into individual decoder stages, selected combinations of stages, and all decoder stages. This experiment examines whether adaptive scale selection should operate throughout coarse-to-fine reconstruction rather than only at a single resolution.

We further compare three LOD prediction strategies in Table~\ref{tb:ablation_stage_lod}, a shared LOD estimator applied independently at each stage, and the proposed stage-conditioned estimators. The latter explicitly condition scale selection on the current decoder state $D_s$, allowing the representation requirement to evolve during reconstruction.

\begin{table}
\centering
\caption{Ablation study of LOD calibration strategies.}
\label{tb:ablation_msfc_main}
\begin{tabular}{l|c}
\hline
Calibration Strategy & PSNR $\uparrow$ \\
\hline\hline
No calibration & 31.52  \\
Minimum reconstruction error & 31.57  \\
Soft reconstruction-error target & 31.62  \\
MSFC without scale penalty &31.61  \\
MSFC & 31.70  \\
\hline
\end{tabular}
\end{table}

\begin{table}
\centering
\caption{Effect of spatial regularization on the learned LOD field.}
\label{tb:ablation_structure_main}
\begin{tabular}{l|c}
\hline
LOD Regularization & PSNR $\uparrow$  \\
\hline\hline
None & 31.49  \\
Uniform smoothness & 31.58 \\
Structure-aware regularization & 31.70 \\
\hline
\end{tabular}
\end{table}

\subsubsection{Effect of LOD Calibration and Structural Regularization}
\label{sec:ablation_regularization}

We finally investigate whether explicit constraints are necessary for learning a meaningful LOD field. Without additional calibration, reconstruction supervision alone may allow different scale assignments to achieve similar errors, causing the predicted LOD to collapse toward a preferred level or behave similarly to unconstrained scale gating.
Table~\ref{tb:ablation_msfc_main} compares the proposed MSFC with variants using no calibration, minimum reconstruction-error supervision, and soft reconstruction-error supervision without the minimal-footprint penalty.

We additionally evaluate the role of structure-aware regularization in Table~\ref{tb:ablation_structure_main}. Uniform smoothness encourages neighboring locations to share similar scales regardless of image structure, whereas the proposed formulation relaxes this constraint around structural boundaries.

\end{document}